\documentclass[runningheads]{llncs}

\usepackage{eccv}

\usepackage{eccvabbrv}

\usepackage{graphicx}
\usepackage[table]{xcolor}
\usepackage{booktabs}

\usepackage[accsupp]{axessibility}  
\definecolor{iccvblue}{rgb}{0.21,0.49,0.74}

\usepackage{hyperref}

\usepackage{orcidlink}
\usepackage{multirow}
\usepackage{marvosym}

\begin{document}

\title{GMODiff: One-Step Gain Map Refinement with Diffusion Priors for Efficient HDR Reconstruction} 

\titlerunning{GMODiff}

\author{Tao Hu\inst{1} \and
Weiyu Zhou\inst{1}  \and
Yanjie Tu\inst{1} \and
Wei Dong\inst{3} \and
Peng Wu\inst{1} \and \\
Qingsen Yan\inst{1,2} \thanks{Corresponding author.} \and
Yanning Zhang\inst{1}}

\authorrunning{T.~Hu et al.}

\institute{Northwestern Polytechnical University, Xi'an 710072, China \and
The Shenzhen Research Institute of Northwestern Polytechnical University, Shenzhen 518057, China
\\ \and
Xi'an University of Architecture and Technology, Xi'an 710055, China\\
\email{taohu@mail.nwpu.edu.cn, qingsenyan@nwpu.edu.cn}}

\maketitle

\begin{abstract}
Pre-trained Latent Diffusion Models (LDMs) have recently shown strong perceptual priors for low-level vision tasks, making them a promising direction for multi-exposure High Dynamic Range (HDR) reconstruction. However, directly applying LDMs to HDR remains challenging due to: (1) limited dynamic-range representation caused by 8-bit latent compression, (2) high inference cost from multi-step denoising, and (3) content hallucination inherent to generative nature.
To address these challenges, we introduce GMODiff, a \textbf{g}ain \textbf{m}ap-driven \textbf{o}ne-step \textbf{diff}usion framework for multi-exposure HDR reconstruction.
Instead of reconstructing full HDR content, we reformulate HDR reconstruction as a degradation-aware Gain Map (GM) refinement problem, where the GM encodes the extended dynamic range while retaining the same bit depth as LDR images.
We initialize the denoising process from an informative regression-based estimate rather than pure noise, allowing the model to generate high-quality GMs in a single denoising step. Furthermore, recognizing that regression-based models excel in content fidelity while LDMs favor perceptual quality, we leverage regression priors to guide both the denoising process and latent decoding of the LDM, suppressing hallucinations while preserving structural accuracy.
Extensive experiments demonstrate that our GMODiff performs favorably against several state-of-the-art methods and is 100$\times$ faster than previous LDM-based methods. The code is available at \href{https://github.com/gbymat/GMODiff}{https://github.com/gbymat/GMODiff}.
\end{abstract}

\section{Introduction}
\begin{figure}[tb]
\centering
\includegraphics[width=1\linewidth]{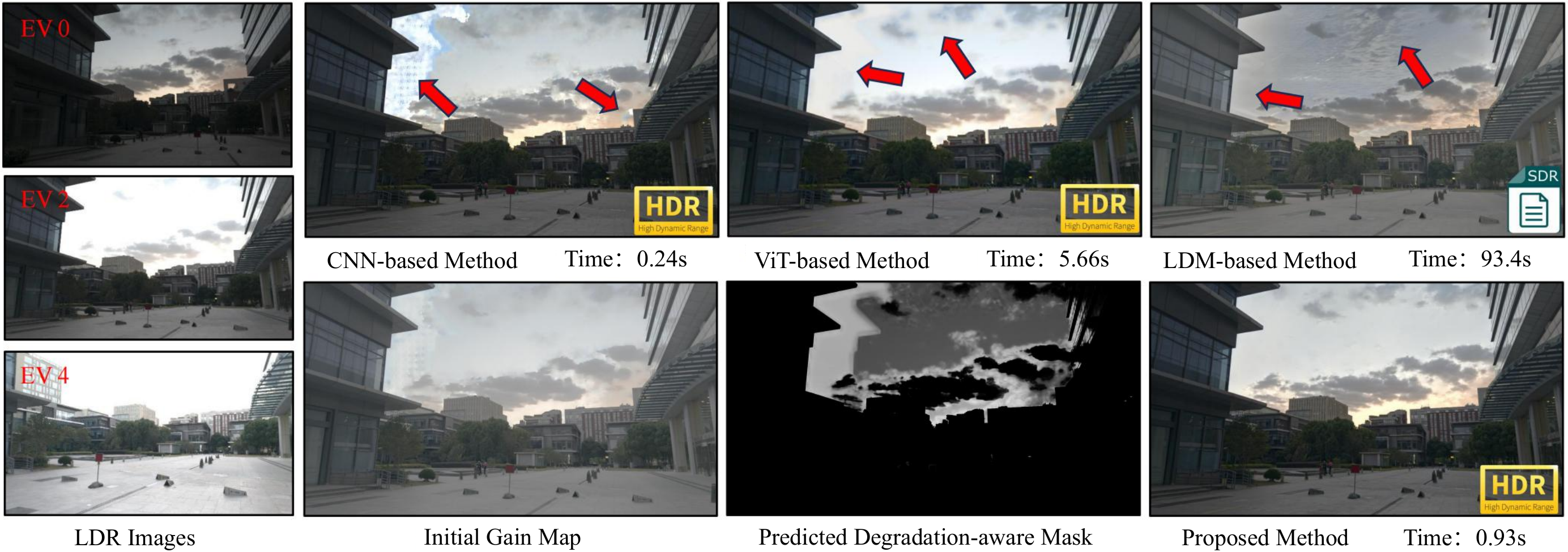}
\caption{
Compared with CNN-based~\cite{yan2019attention} and ViT-based~\cite{tel2023alignment} methods, the LDM-based method UltraFusion~\cite{chen2025ultrafusion} generates perceptually compelling HDR-like images, highlighting the potential of LDMs for HDR reconstruction. However, it incurs high computational cost and may introduce hallucination artifacts that compromise physical fidelity.
}
\label{ins}
\end{figure}
Multi-exposure High Dynamic Range (HDR) imaging fuses Low Dynamic Range (LDR) images of varying exposures to recover scene radiance, but suffers from motion-induced misalignment in dynamic scenes.
Recent Deep Neural Networks (DNNs), including Convolutional Neural Networks (CNNs) \cite{yan2019attention,wu2018deep,yan2020deep} and Vision Transformers (ViTs) \cite{liu2022ghost,yan2023unified,tel2023alignment}, have greatly improved HDR reconstruction in dynamic scenes. 
Despite their advancements, these methods simply learn a mapping network between LDR-HDR paired data using pixel-wise losses \cite{wang2021deep}, which often yields perceptually unsatisfactory results. This limitation is particularly pronounced when essential content is missing in severely overexposed regions due to object or camera motion.

Recent advances in Latent Diffusion models (LDMs) \cite{ho2020denoising,dhariwal2021diffusion}, particularly large-scale pre-trained text-to-image (T2I) models ~\cite{rombach2022high,saharia2022photorealistic} trained on billions of image-text pairs, possess powerful natural image priors. These priors make LDMs a promising solution for addressing the perceptual shortcomings of current HDR reconstruction methods, especially in mitigating motion-induced ghosting and other complex artifacts. 
However, directly applying LDMs to high-quality HDR image reconstruction remains non-trivial due to three key challenges:
\textbf{(1) Limited Dynamic Range Representation:}  The autoencoders of pre-trained LDMs compress linear HDR distributions into an 8-bit perceptual latent space, limiting their capacity to represent the dynamic range of HDR content \cite{wang2025lediff,guan2025hdr}.
\textbf{(2) High Inference Cost:} LDMs demand a substantial number of iteration steps in the denoising process, which is time-consuming even with a high-end GPU card \cite{yan2023toward,hu2024generating}. 
\textbf{(3) Content Hallucination:} The generative nature of LDMs can lead to hallucinated content that deviates from realistic scenes.
As shown in Fig.~\ref{ins}, UltraFusion \cite{chen2025ultrafusion} produces human-perceptually consistent HDR-like content with a compressed dynamic range but suffer from high latency and hallucination.

These limitations motivate a paradigm shift in leveraging LDMs for HDR reconstruction. Our work draws from three key insights: 
\textbf{First,} inspired by the observation that an HDR image can be decomposed into a LDR component and a 8 bit-depth Gain Map (GM) \cite{apple2021edr,adobe2024gainmap,google2024ultrahdr} that encodes the extended dynamic range, we reframe HDR reconstruction as a conditionally guided GM refinement task to fully exploit pre-trained LDM priors without domain mismatches.
\textbf{Second,} since LDR inputs already provide rich structural and semantic cues, we initialize the denoising process from an informative prior (\eg, a regression-based GM estimate) rather than pure noise, significantly accelerating inference. 
\textbf{Third,} recognizing that regression-based models excel in content fidelity while LDMs favor perceptual quality, we leverage regression-based priors to guide the LDM’s denoising and latent decoding processes.

Building on these principles, we propose GMODiff, a gain map-driven one-step diffusion model for multi-exposure HDR reconstruction. To the best of our knowledge, GMODiff is the first LDM-based method designed for multi-exposure HDR reconstruction. Rather than training a LDM to directly reconstruct full HDR radiance, we adopt a two-stage fine-tuning strategy to adapt a pre-trained multi-step LDM for single-step GM refinement.
Specifically, in the first stage, we train a Degradation-aware Regressor Network (DaReg) via a dual-learning strategy to produce an initial GM and extract spatial embeddings that highlight regions where the estimate is unreliable. This learning scheme provides the LDM with an informative prior for initialization and directs its attention to ambiguous or error-prone areas, significantly accelerating convergence and improving structural fidelity.
In the second stage, we fine-tune the LDM to perform one-step denoising conditioned on these regression-based priors, enabling accurate GM correction and high-fidelity HDR reconstruction at dramatically reduced computational cost.
In summary, our main contributions are threefold:
\begin{itemize}
    \item We propose GMODiff, a one-step diffusion framework that reframes HDR reconstruction as a conditionally guided gain map refinement task, fully leveraging pre-trained LDMs without redesigning the latent space.
    
    \item  We design a degradation-aware refinement pipeline that seamlessly integrates regression-based fidelity with diffusion-driven perceptual enhancement: the DaReg produces an initial gain map and degradation-aware embeddings, which guide both the denoising process and the latent decoding of the LDM, effectively suppressing hallucinations while preserving structural accuracy.
    
    \item Extensive experiments demonstrate that the proposed GMODiff delivers superior visual quality with significantly reduced ghosting and artifacts, while achieving high inference efficiency.
\end{itemize} 

\section{Related Work}
\subsection{Multi-exposure HDR Reconstruction}
Multi-exposure HDR imaging has advanced significantly, with methods broadly divided into traditional and deep learning-based categories. Traditional methods address motion artifacts via motion rejection~\cite{Gallo2009Artifact}, registration~\cite{bogoni2000extending, kang2003high}, and patch matching~\cite{sen2012robust,Hu2013deghosting}, but their effectiveness hinges on preprocessing accuracy, often failing with large/complex motion. 
In contrast, state-of-the-art HDR reconstruction methods now predominantly leverage the powerful nonlinear modeling capabilities of DNNs.
Recent advances incorporate attention mechanisms~\cite{yan2023unified,yan2019attention,11433535} and ViT architectures~\cite{tel2023alignment,yan2023unified,liu2022ghost,zhou2026high} to improve feature correspondence and capture long-range dependencies. Meanwhile, optical flow–guided approaches~\cite{kalantari2017deep,kong2024safnet} explicitly address misalignment by warping input frames according to estimated motion fields.
Moreover, generative models, including Generative Adversarial Networks (GANs)~\cite{niu2021hdr} and diffusion models~\cite{yan2023toward,hu2024generating,chen2025ultrafusion,wang2025lediff,yan2024dynamic}, have been successfully adapted to synthesize photorealistic HDR content, even from imperfect or severely degraded inputs. 
Additional innovative frameworks, such as \cite{li2025afunet,ni2025semantic,11353407}, have further expanded the scope of DNN applications in this field, introducing novel architectures and training strategies to improve reconstruction quality.   
Despite these advances, most existing methods either learn a direct LDR-to-HDR mapping using pixel-wise losses, or train generative models from scratch on limited datasets. These often yield unsatisfactory results, especially when critical content is missing in severely over/underexposed regions due to motion.

\subsection{Gain Map-based HDR Reconstruction}
To support consistent and device-adaptive HDR image rendering, several recent industry standards have introduced a dual-layer image representation \cite{apple2021edr,adobe2024gainmap,google2024ultrahdr}. This format jointly stores a standard LDR image together with an auxiliary gain map, enabling flexible reconstruction of HDR content across diverse display environments.
Recently, several GM-based HDR reconstruction methods have emerged. Liao \etal \cite{liao2025learning} introduced the Gain Map–based Inverse Tone Mapping (GM-ITM) task, shifting the learning objective from direct HDR prediction to gain map estimation, thereby achieving greater flexibility.
Canham \etal \cite{canham2024gamma} proposed an alternative to the multiplicative GM by introducing a pixel-wise exponent map, and in their subsequent work, they further introduced an MLP-based GM encoder \cite{canham2025gain}, which demonstrated improved efficiency in HDR recovery.
More recently, Guan \etal \cite{guan2025hdr} leveraged two cascaded diffusion models to separately synthesize the LDR image and its corresponding GM for HDR image generation.
Compared to traditional HDR reconstruction methods, the GM exhibits a more balanced distribution \cite{liao2025learning} and offers greater flexibility during the reconstruction process. Motivated by these benefits, we reformulated the multi-exposure HDR reconstruction task as a conditionally guided GM estimation problem, which allows us to fully exploit pre-trained LDM priors without suffering from domain mismatches.

\subsection{Diffusion Models}
Diffusion models \cite{rombach2022high,saharia2022photorealistic} have emerged as a powerful class of generative frameworks, capable of synthesizing high-fidelity data through iterative denoising of random noise. They have already demonstrated impressive performance across a wide range of low-level vision tasks-such as super-resolution~\cite{wu2024one}, denoising~\cite{du2025superpc}, dehazing~\cite{lan2025exploiting,lan2025schrodinger}, JPEG artifact removal~\cite{guo2025compression}, image enhancement~\cite{10916787},  and deblurring~\cite{xie2025diffusion}, thanks to their ability to model fine-grained details and generate visually compelling outputs, often surpassing traditional image restoration methods.
Despite their success, the practical deployment of diffusion models remains hindered by their high computational cost. To mitigate this bottleneck, recent efforts \cite{lu2022dpm,song2020denoising} have focused on accelerating diffusion by reducing the number of denoising steps. However, aggressive step reduction typically leads to a noticeable degradation in output quality.
Recently, img2img \cite{parmar2024one} proposed a general framework that adapts pre-trained single-step diffusion models to new tasks and domains via adversarial learning. Meanwhile, several recent studies~\cite{guo2025compression,wu2024one,du2025superpc} have explored single-step diffusion models for low-level vision tasks. These methods enable efficient inference while retaining the rich generative priors embedded in large-scale pre-trained diffusion models. 
Nevertheless, directly applying this efficient diffusion paradigm to HDR generation remains challenging. Most pre-trained diffusion models rely on autoencoders trained on 8-bit LDR images, which compress HDR’s high bit-depth into a perceptually encoded, low-dimensional latent space, severely limiting their ability to capture the full dynamic range of the real world.

\begin{figure}[t]
\centering
\includegraphics[width=1\linewidth]{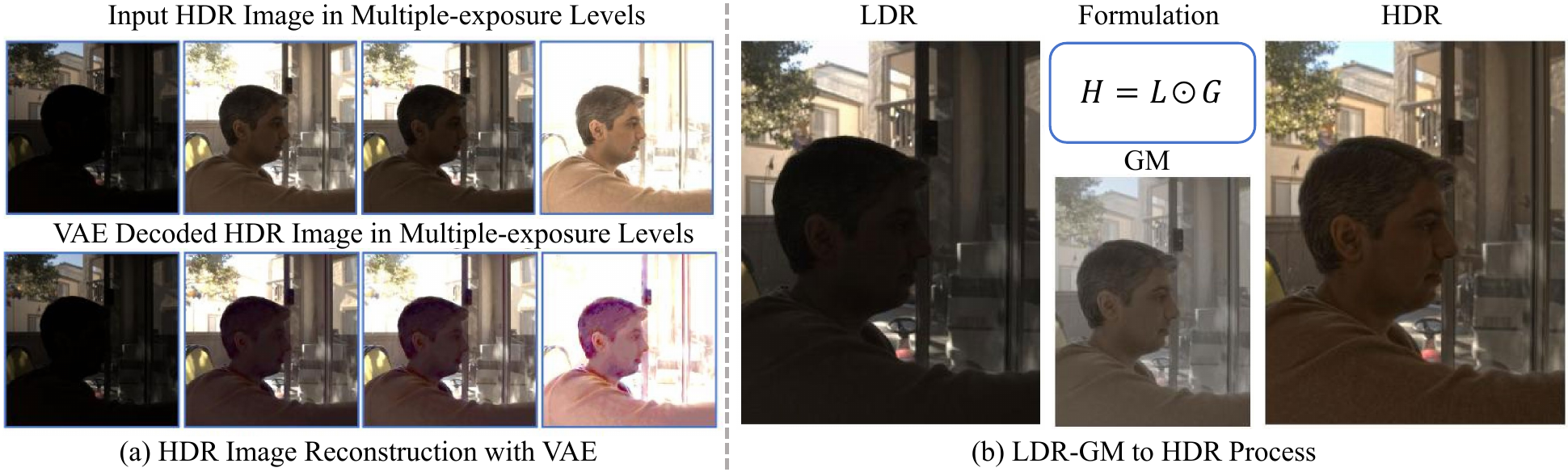}
\caption{
(a) Limitations of the vanilla LDMs in generating HDR content. The limitation of the LDM VAE in encoding and decoding an HDR image, visualized in multiple exposure levels, which reveals a significant fidelity loss, especially in the shadow. (b) The GM encodes pixel-wise dynamic range adjustments for the LDR image and can be used to reconstruct the HDR image via element-wise multiplication.
}
\label{vaedecoder}
\end{figure}

\section{Methodology}
\begin{figure*}[t]
\centering
\includegraphics[width=1\textwidth]{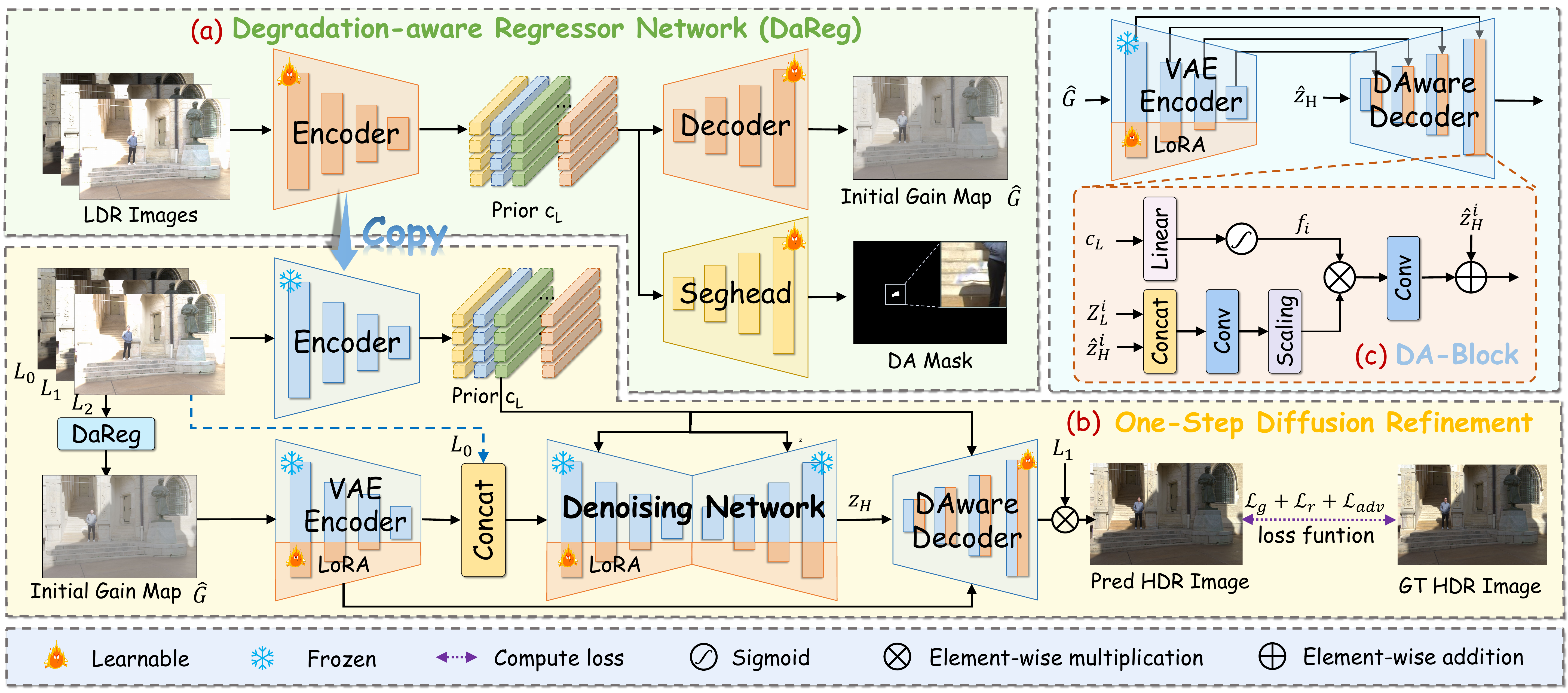}
\caption{Overview of the proposed GMODiff framework.  
(a) We train a DaReg via a dual-learning strategy to produce two regression-based priors: an initial gain map $\hat{G}$ and spatial embeddings $c_L$. The $c_L$ implicitly encode regions where the $\hat{G}$ is unreliable, serving as a degradation-aware guidance prior.  
(b) One-Step Diffusion Refinement initializes the denoising process from $\hat{G}$ and performs single-step denoising conditioned on $c_L$ to generate a high-fidelity GM latent code $Z_H$.  
(c) The DA Decoder leverages encoder features from the initial gain map $\hat{G}$, guided by $c_L$, to recover fine image details while avoiding the introduction of artifacts inherent in $\hat{G}$.}
\label{frame}
\end{figure*}

As shown in Fig.~\ref{frame}, GMODiff is trained in two stages.
During  the first stage (Sec.~\ref{secdareg}), we pretrain a Degradation-aware Regressor Network (DaReg) to predict an initial gain map and degradation-aware embeddings from the LDR inputs.
During  the second stage (Sec.~\ref{secdm}), we fine-tune a pre-trained LDM using these regression priors to refine the initial gain map and reconstruct a high-quality HDR image.
\subsection{Preliminaries}
As shown in Fig.~\ref{vaedecoder} (a), while the variational autoencoder (VAE) of a LDM can encode visually plausible LDR representations that resemble HDR content under standard viewing conditions, it fundamentally fails to preserve true high dynamic range-particularly in highlights and shadows. Since the denoising network operates in an 8-bit perceptual latent space, a naive solution to enable HDR generation would be to fine-tune both the VAE and the denoising network on HDR data. However, this approach demands large-scale HDR training sets and discards the rich generative priors learned from massive LDR images \cite{wang2025lediff}.

Recent advances in display technology have popularized dual-layer HDR formats \cite{apple2021edr,google2024ultrahdr}, which decompose an HDR image $\mathbf{H}$ into an LDR base layer $\mathbf{L}$ and an 8-bit gain map $\mathbf{G}$.
The gain map is obtained by pixel-wise division of $\mathbf{H}$ by $\mathbf{L}$, followed by logarithmic compression.
This representation has also shown promise for diffusion-based HDR generation. Specifically, Guan \etal \cite{guan2025hdr} employed two cascaded diffusion models to separately synthesize the LDR base image and its corresponding gain map for HDR image generation.
As illustrated in Fig.~\ref{vaedecoder} (b), the HDR image can be recovered from the dual-layer representation as:
\begin{equation}
\mathbf{H} = (\mathbf{L} + \alpha) \cdot \exp_2\big(1 + \mathbf{G} \cdot Q_{\max}\big),
\label{eq:gm_gt}
\end{equation}
where $\mathbf{G}\in[0,1]$ is the normalized gain map, $\exp_2(\cdot)$ denotes the inverse log transform, $Q_{\max}$ corresponds to the maximum gain value, and $\alpha=1/64$ is a scaling constant that preserves detail in near-black regions. Crucially, both the LDR image and gain map are represented at 8-bit depth, allowing us to directly exploit pre-trained LDMs without VAE retraining or latent-space redesign.

\subsection{Degradation-aware Regressor Network}
\label{secdareg}
The first stage produces two regression-based priors: an initial gain map $\hat{\mathbf{G}}$ and multi-scale spatial embeddings $\mathbf{c}_L$. These priors are used to initialize and condition the one-step diffusion refinement, ensuring high structural fidelity while enabling perceptual enhancement.

\noindent\textbf{Overall Architecture.} 
Our DaReg adopts a NAFNet backbone~\cite{chen2022simple} augmented with a lightweight degradation-aware segmentation head.  
Given the input LDR sequence $\{L_1, L_2, \ldots, L_N\}$, we first apply gamma correction to each $L_i$ to obtain its HDR version $H_i$, and form a six-channel tensor $X_i = [L_i, H_i]$ by channel-wise concatenation.  
Following~\cite{yan2019attention,liu2022ghost}, an implicit alignment module processes the LDR inputs and fuses multi-exposure features before feeding them into the network.  
As shown in Fig.~\ref{frame}, DaReg encodes the fused features into a latent representation $\mathbf{c}_L$. For diffusion-based refinement, the spatial feature map is flattened and projected into a sequence of spatial embeddings:
\begin{equation}
\mathbf{c}_L = \left\{ \mathbf{c}_{L_k} \in \mathbb{R}^d \right\}_{k=1}^K,
\end{equation}
where $K$ denotes the number of spatial locations, and each $\mathbf{c}_{L_k}$ corresponds to the feature representation at one spatial location. For notational simplicity, we use $\mathbf{c}_L$ to denote this latent representation in both its spatial form for initial gain-map reconstruction and its tokenized form for subsequent diffusion refinement.
The decoder reconstructs the initial gain map using the spatial form of $\mathbf{c}_L$, while the segmentation head predicts a pixel-wise reliability map for degraded regions (\eg, due to motion or saturation). After the first stage of training, the predicted reliability map is fused with the tokenized $\mathbf{c}_L$ to provide explicit conditional cues for uncertain regions during subsequent refinement.

\noindent\textbf{Initial GM Reconstruction.}
Given an input LDR sequence, the encoder extracts hierarchical features, where each stage consists of multiple Nonlinear Activation-Free Blocks~\cite{chen2022simple}. A symmetric decoder then reconstructs the initial gain map $\hat{\mathbf{G}}$ from the spatial form of $\mathbf{c}_L$ as $\hat{\mathbf{G}} = D\left(\mathbf{c}_L; \theta_D\right)$, where each decoder stage receives skip-connected features from the corresponding encoder stage to progressively reconstruct the initial GM.

During training, we compute the ground-truth gain map $\mathbf{G}$ using Eq.~\eqref{eq:gm_gt}, where $\mathbf{L}$ is the reference frame of the LDR sequence. 
To enable direct supervision with HDR images and ensure differentiability at zero, we replace the logarithmic compression with the $\mu$-law function. Accordingly, the $\exp_2(\cdot)$ in Eq.~\eqref{eq:gm_gt} is substituted by the inverse $\mu$-law mapping: $R(x) = \frac{e^{x \cdot \log(1+\mu)} - 1}{\mu}$, with $\mu$=100.
We supervise the initial gain map prediction $\hat{\mathbf{G}}$ via an $\mathcal{L}_1$ regression loss:
\begin{equation}
    \mathcal{L}_{\text{gm}} = \|\hat{\mathbf{G}} - \mathbf{G}\|_1.
\end{equation}
Additionally, we enforce consistency between the reconstructed HDR image $\hat{\mathbf{H}}$ (from $\hat{\mathbf{G}}$) and the ground-truth $\mathbf{H}$:
\begin{equation}
\label{eq:lr}
\mathcal{L}_r = \left\|\mathcal{T}(\mathbf{H}) - \mathcal{T}(\hat{\mathbf{H}})\right\|_1 + \lambda \left\|\phi_{i, j}(\mathcal{T}(\mathbf{H})) - \phi_{i, j}(\mathcal{T}(\hat{\mathbf{H}}))\right\|_1,
\end{equation}
where $\mathcal{T}(\cdot)$ denotes the $\mu$-law function, $\phi_{i,j}(\cdot)$ is the $j$-th feature map of VGG19 after the $i$-th max-pooling layer, and $\lambda = 1\mathrm{e}{-2}$ is a weighting hyperparameter.

\noindent\textbf{Degradation-Aware Embedding.}
By incorporating a reconstruction objective, DaReg implicitly learns regression-based priors that facilitate dynamic range expansion. To better guide the LDM in refining the initial GM, we additionally introduce a segmentation head to predict a degradation-aware mask, which allows the refinement process to focus on uncertain regions and avoids hallucinating implausible details in well-reconstructed areas.

Specifically, given an input LDR sequence, we extract multi-scale latent representations with the DaReg encoder as before. For the degradation-aware mask \(M\), we process the encoder features to the same number of channels using \(1\times1\) convolutions, and upsample them to the original resolution for concatenation to produce \(\bar{\mathbf{c}}_L\). The segmentation head \(h(\cdot; \phi)\) processes \(\bar{\mathbf{c}}_L\) through convolutional layers with 64 hidden channels, followed by a sigmoid activation, and outputs a predict mask: $\hat{M} = h(\bar{\mathbf{c}}_L; \phi),$ where \(\hat{M} \in [0,1]\) denotes the predicted degradation probability at each pixel $p$. Notably, to prevent the gradient of the segmentation head from affecting the reconstruction accuracy, we perform a detach operation before feature processing. The obtained degradation-aware mask is then compressed through a lightweight CNN network in the next training stage and fused with the original \(\mathbf{c}_L\) to serve as the conditional guidance for the LDMs.

We obtain the ground-truth mask $M$ using a simple yet effective pixel-consistency criterion:
\begin{equation}
\label{eqn:reliability_mask}
    M(p) = 
    \begin{cases}
        1 & \text{if } \big\| \mathcal{T}(\mathbf{H}(p)) - \mathcal{T}(\hat{\mathbf{H}}(p)) \big\|_1 > \sigma, \\
        0 & \text{otherwise},
    \end{cases}
\end{equation}
where $p$ denotes a pixel location, $\|\cdot\|_1$ averages the absolute difference across RGB channels to yield a scalar, and $\sigma = 4/255$ is a fixed threshold. This formulation is motivated by the observation that static regions yield consistent reconstructions-even with simple image blending-whereas motion or overexposure causes significant pixel-level discrepancies.
The loss function adopted is the binary cross-entropy (BCE) loss between $\hat{M}$ and ${M}$.

\subsection{Prior-Guided One-Step Diffusion Refinement}
\label{secdm}
In the second stage, our goal is to efficiently leverage the strong image priors of pre-trained LDMs to address degradations in the initial GM (\eg, ghosting and artifacts), thereby reconstructing high-quality HDR images.

\noindent\textbf{One-step Denoising Process.}
Based on the prior analysis, we formulate GM refinement as a conditionally guided image-to-image denoising process, which enables the adoption of an efficient one-step denoising strategy \cite{wu2024one} and accelerates convergence. 
Our GMODiff first encodes the initial GM into the latent space via an encoder \(E_\theta\), resulting in \(\mathbf{Z}_L = E_\theta(\hat{\mathbf{G}})\). A single-step denoising operation $F(\mathbf{Z}_L;\mathbf{c})$ is then performed to predict the noise \(\hat{\epsilon}\), which enables us to derive the estimated high-quality latent representation \(\hat{\mathbf{Z}}_H\):
\begin{equation}
\hat{\mathbf{Z}}_H=F(\mathbf{Z}_L;\mathbf{c})\triangleq\frac{\mathbf{Z}_L-\sqrt{1-\bar{\alpha}_{T_L}} \varepsilon_\theta\left(\mathbf{Z}_L ; \mathbf{c}, T\right)}{\sqrt{\bar{\alpha}_{T}}},
\end{equation}
where \(\varepsilon_\theta\) denotes the denoising network, $c$ is the guidance condition, \(T \in [0, T]\) is the predefined diffusion timestep, and \(\bar{\alpha}\) represents the predefined noise schedule. 
Specifically, we append trainable LoRA \cite{hu2022lora} layers to the encoder \(E_\phi\) and the diffusion network \(\epsilon_\phi\), fine-tuning them into \(E_\theta\) and \(\epsilon_\theta\) using our training data. The condition \(\mathbf{c}\) includes the regression prior \(\mathbf{c}_L\) and the underexposed LDR image \(L_0\).
$\mathbf{c}_L$ is injected into the hidden layers of the denoising network, whereas $L_0$ is first encoded into a feature map, then concatenated with $\mathbf{Z}_L$ and fed into the denoising network, with the number of channels in the first convolutional layer adjusted accordingly.

\noindent\textbf{Degradation-aware Decoder.}
The pre-trained LDM performs denoising in a highly compressed latent space (\eg,\(\mathcal{R} \in \mathbb{R}^{4 \times \frac{H}{8} \times \frac{W}{8}}\)), which can result in detail distortion in generated content \cite{parmar2024one}. 
To address this issue, we propose a Degradation-aware Decoder that leverages the encoder features of $\hat{\mathbf{G}}$ under the guidance of degradation-aware priors to facilitate refinement of detail while suppressing artifacts inherent in the initial GM $\hat{\mathbf{G}}$.

Specifically, we obtain multi-scale features $\mathbf{Z}_L^i$ of the initial GM $\hat{\mathbf{G}}$ via the VAE encoder through forward propagation, while $\mathbf{\hat{Z}}_H$ is derived from a one-step diffusion process. Although the initial gain map, reconstructed via a regression approach, endows $\mathbf{Z}_L^i$ with rich detail fidelity, it may introduce noticeable artifacts. To address this, we propose a Degradation-aware Refinement Block guided by the embedding $\mathbf{c}_L$, which leverages degradation-aware embeddings to identify unreliable regions in $\mathbf{Z}_L^i$.
As illustrated in Fig.~\ref{frame} (c), the Degradation-aware Refinement Block first concatenates the decoder intermediate features $\mathbf{\hat{Z}}_H^i$ with $\mathbf{Z}_L^i$ and applies a $3\times3$ convolution to obtain refined features $\hat{\mathbf{Z}}^i$:
\begin{equation}
    \hat{\mathbf{Z}}^i = \operatorname{Conv}_{3\times3}\big( [\mathbf{\hat{Z}}_H^i, \mathbf{Z}_L^i] \big),
\end{equation}
where $[\cdot]$ denotes channel-wise concatenation.
The $\mathbf{c}_L$ is then passed through an MLP followed by a sigmoid activation to produce a channel-wise modulation vector:
\begin{equation}
\begin{aligned}
    & \mathbf{f}_i = \operatorname{Sigmoid}\big(\operatorname{MLP}(\mathbf{c}_L)\big), \\
    & \mathbf{\hat{Z}}_H^{i+1} = \operatorname{Conv}_{3\times3}\big( \hat{\mathbf{Z}}^i \odot \mathbf{f}_i \big) + \mathbf{\hat{Z}}_H^i,
\end{aligned}
\end{equation}
where $\odot$ denotes channel-wise multiplication, and $\mathbf{\hat{Z}}_H^{i+1}$ represents the updated decoder feature after detail modulation and residual connection.

\noindent\textbf{Training Strategy}
During the training phase, the loss function comprises the GM reconstruction loss and HDR reconstruction loss. Following \cite{guo2025compression}, we also compute the adversarial loss \(\mathcal{L}_{\text{adv}}\) between the predicted latent gain map $\mathbf{\hat{Z}}_H$ and the GT latent gain map $\mathbf{{Z}}_H$, where the discriminator consists of a frozen pre-trained denoising UNet encoder and two trainable MLP layers. The total objective for decoder training is:
\begin{equation}
    \mathcal{L}_{diff} =  \mathcal{L}_{gm} + \mathcal{L}_{r} + \lambda\mathcal{L}_{adv}, 
\end{equation}
where \(\lambda\) denotes the balancing parameter, which is empirically set to 0.001.

Additionally, considering that the initial GM $\hat{\mathbf{G}}$ obtained in the first stage may overfit to the training set, which impairs GMODiff’s ability to handle its degradations,  we propose a group training strategy for GMODiff to learn robust degradation mitigation.
Specifically, we divide the training set into N groups and train \(N+1\) DaRegs accordingly, with the extra model trained on the entire training dataset. During GMODiff training, we randomly select one DaReg to provide $\hat{\mathbf{G}}$ and \(\mathbf{c}_L\). The training objective for GMODiff is ultimately defined as the expected loss over such a random selections:
\begin{equation}
\label{eq:group_training}
\mathcal{L} = \mathbb{E}{_k \sim \text{Uniform}(0, N)} \left[ \mathcal{L}_{diff} \left( \hat{G}_k, \mathbf{c}_{L,k}; H ; G\right) \right],
\end{equation}
where $k$ is the index of a randomly selection.

\section{Experiments}

\begin{table*}[t]
\caption{Quantitative comparison on the in-distribution test split of our collected  dataset. Best and second-best results per metric are highlighted in \textcolor{red}{red} and \textcolor{iccvblue}{blue}, respectively.}
\Large
\centering
\resizebox{\textwidth}{!}{
\begin{tabular}{c|cccccc|cccc}
\toprule[0.15em]
Metric & DeepHDR \cite{wu2018deep} & AHDRNet \cite{yan2019attention} & SAFNet \cite{niu2021hdr} & HDR-Trans \cite{liu2022ghost} & SCTNet \cite{tel2023alignment} & AFUNet \cite{li2025afunet} & LE-Diff \cite{wang2025lediff} & GMDD \cite{guan2025hdr} & DiffHDR \cite{yan2023toward} & Ours \\
\midrule[0.15em]
PU21-PSNR$\uparrow$      & 32.91 & 35.37 & 36.53 & 36.30 & 35.47 & \textcolor{red}{36.91} & - & 16.31 & 35.43 & \textcolor{iccvblue}{36.68} \\
PU21-SSIM$\uparrow$      & 0.9814 & 0.9861 & 0.9876 & {0.9868} & 0.9871 & \textcolor{red}{0.9889} & - & 0.7740 & 0.9861 & \textcolor{iccvblue}{0.9882} \\
HDR-VDP2$\uparrow$   & 67.98 & 68.74 & 68.84 & 68.16 & 68.92 & \textcolor{red}{70.77} & - & 63.01 & 68.79 & \textcolor{red}{70.77} \\
DISTS$\downarrow$   & 0.0221 & 0.0172 & \textcolor{iccvblue}{0.0112} & 0.0140 & 0.0121 & 0.0120 & - & 0.0856 & 0.0125 & \textcolor{red}{0.0103} \\ 
\midrule
MANIQA$\uparrow$    & 0.5786 & 0.5828 & 0.5817 & 0.5804 & \textcolor{iccvblue}{0.5857} & 0.5820 & 0.5621 & 0.5738 & 0.5834 & \textcolor{red}{0.5881} \\
MUSIQ$\uparrow$     & 59.59 & 59.77 & 60.59 & 60.14 & {60.59} & \textcolor{iccvblue}{60.66} & 57.07 & 57.88 & {60.39} & \textcolor{red}{60.96} \\
CLIPIQA$\uparrow$   & 0.3946 & 0.4085 & 0.4199 & 0.4105 & 0.4183 & 0.4164 & 0.3675 & \textcolor{iccvblue}{0.4225} & 0.4093 & \textcolor{red}{0.4253} \\
\bottomrule[0.15em]
\end{tabular}}
\label{tab:main_results}
\vspace{0mm}
\end{table*}

\begin{table*}[t]
\caption{Quantitative comparisons on the Prabhakar \etal's \cite{prabhakar2019fast} dataset. Best and second-best results per metric are highlighted in \textcolor{red}{red} and \textcolor{iccvblue}{blue}, respectively.}
\Large
\centering
\vspace{.10mm}
\resizebox{\textwidth}{!}{
\begin{tabular}{c|cccccc|cccc}
\toprule[0.15em]
Metric & DeepHDR \cite{wu2018deep} & AHDRNet \cite{yan2019attention} & SAFNet \cite{niu2021hdr} & HDR-Trans \cite{liu2022ghost} & SCTNet \cite{tel2023alignment} & AFUNet \cite{li2025afunet} & LE-Diff \cite{wang2025lediff} & GMDD \cite{guan2025hdr} & DiffHDR \cite{yan2023toward} & Ours \\
\midrule[0.15em]
PU21-PSNR$\uparrow$      & 23.06 & 26.30 & 26.13 & 25.18 & \textcolor{iccvblue}{27.41} & \textcolor{red}{27.78} & - & 15.03 & 25.31 & 26.86\\ 
PU21-SSIM$\uparrow$      & 0.8775 & 0.9413 & 0.9410 & 0.9401 & \textcolor{red}{0.9454} & 0.9398 & - & 0.7577 & 0.9389 & \textcolor{iccvblue}{0.9419} \\ 
HDR-VDP2$\uparrow$       & 63.79 & \textcolor{iccvblue}{67.67} & 66.59 & 64.95 & 64.78 & 66.46 & - & 63.68 & 66.78 & \textcolor{red}{67.91} \\ 
DISTS$\downarrow$        & 0.0810 & 0.0433 & 0.0521 & 0.0644 & 0.0407 & 0.0589 & - & 0.1383 & \textcolor{iccvblue}{0.0395} & \textcolor{red}{0.0379} \\ 
\midrule
MANIQA$\uparrow$         & 0.5222 & \textcolor{iccvblue}{0.5450} & 0.5410 & 0.5319 & 0.5412 & 0.5305 & 0.4413 & 0.5011 & 0.5369 & \textcolor{red}{0.5469} \\ 
MUSIQ$\uparrow$          & 52.95 & 52.55 & 53.25 & \textcolor{iccvblue}{54.61} & 53.59 & 52.70 & 34.57 & 42.41 & 53.67 & \textcolor{red}{55.48} \\ 
CLIPIQA$\uparrow$        & 0.3938 & 0.4387 & 0.4478 & 0.4489 & \textcolor{iccvblue}{0.4513} & 0.4395 & 0.3767 & 0.3861 & 0.4474 & \textcolor{red}{0.4602} \\ 
\bottomrule[0.15em]
\end{tabular}}
\label{tab:main_results2}
\vspace{0mm}
\end{table*}

\begin{figure*}[tb]
\centering
\includegraphics[width=1\linewidth]{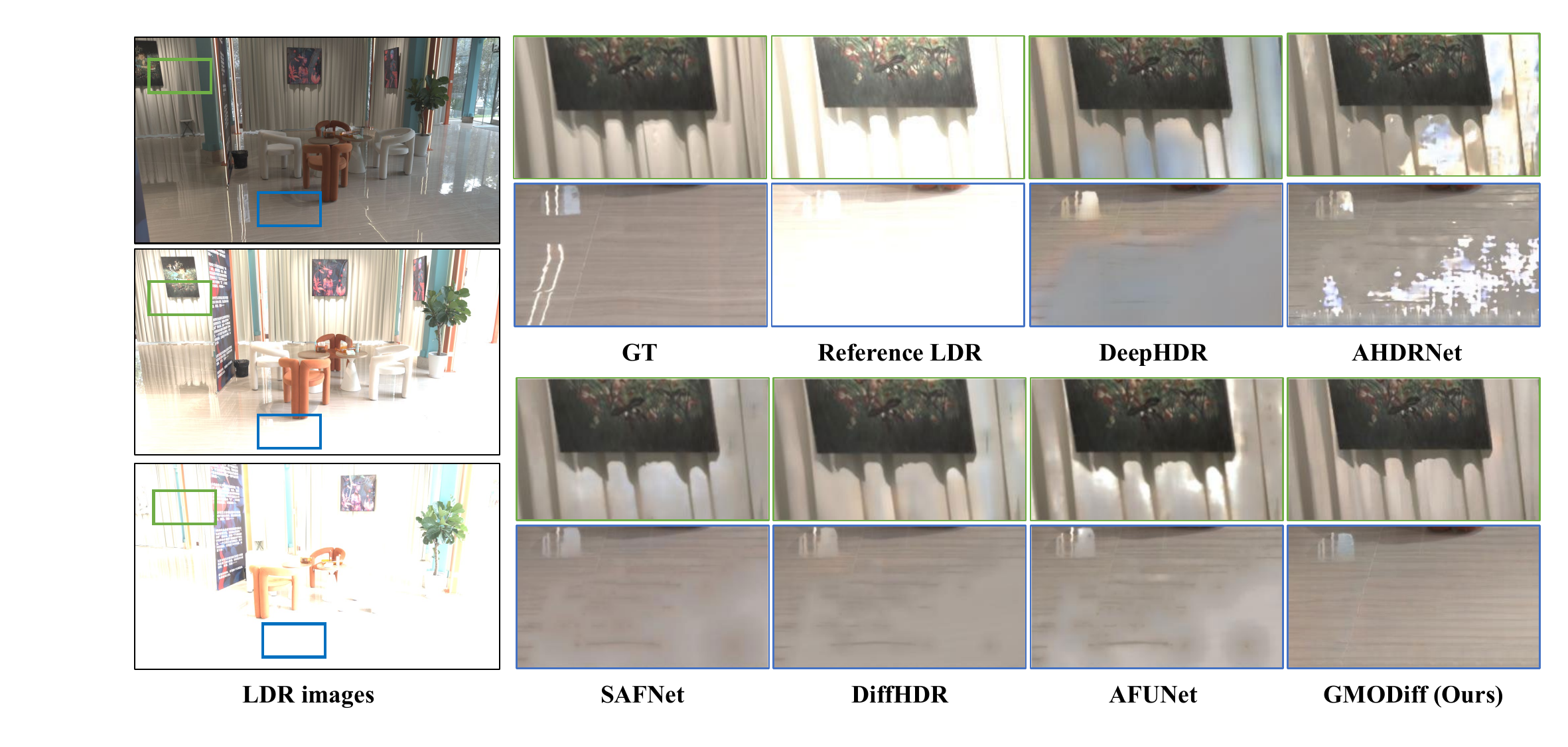}
\caption{Visual comparisons are conducted on the test set, with a focus on zoomed-in local regions of HDR images reconstructed by our method and competing techniques. Compared with existing methods, GMODiff more faithfully recovers local structures and illumination details while preserving visually natural brightness transitions.
}
\label{compare1}
\end{figure*}
\begin{figure*}[tb]
\centering
\includegraphics[width=1\linewidth]{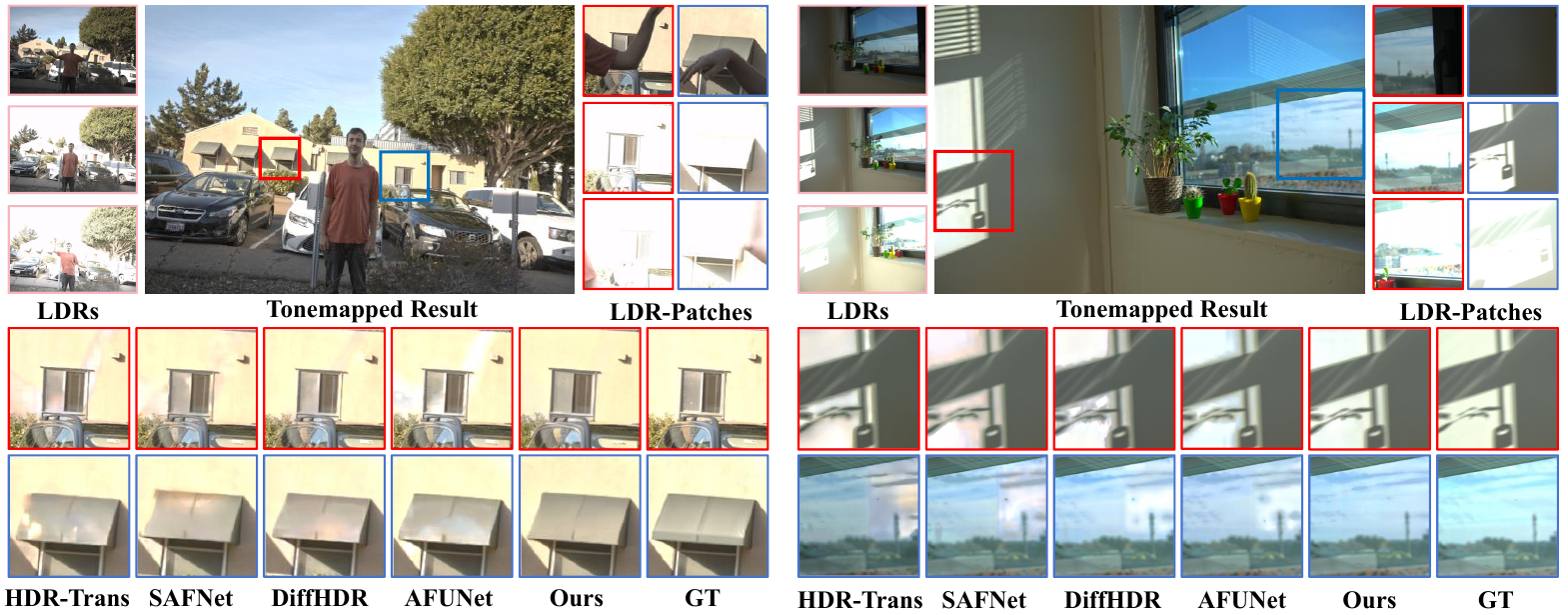}
\caption{Visual comparisons are conducted on testing data , focusing on zoomed-in local areas of the HDR images estimated by our method and the compared techniques. Our model demonstrates the ability to generate HDR images of superior quality.
}
\label{compare2}
\end{figure*}

\subsection{Experimental Settings}
\noindent\textbf{Datasets.} 
Scenes and motion variations within a single dataset are limited. To better validate the model’s generalization ability under diverse real-world conditions, we collect publicly available real-world datasets \cite{kong2024safnet,tel2023alignment,kalantari2017deep} commonly used for multi-exposure HDR reconstruction, which cover a wide range of illumination and motion conditions. We follow the official training and test splits of all datasets. For evaluation, we conduct quantitative comparisons on datasets with ground truth, including the test split of our training data (in-distribution) and the Prabhakar \etal \cite{prabhakar2019fast} dataset (out-of-distribution) for real-world robustness testing. In addition, we use the Sen \etal dataset \cite{sen2012robust} and the Tursun \etal dataset \cite{Tursun2016data} solely for qualitative comparisons, as they do not provide ground-truth HDR images.

\noindent\textbf{Implementation Details.}
We implemented our model in PyTorch and used the \texttt{accelerate} library for mixed-precision training with FP16.
In stage one, we optimize DaReg using the Adam optimizer with a learning rate of 2e-4. The training data is preprocessed by extracting $256\times256$ patches, and we use a batch size of 16. 
DaReg adopts the NAFNet architecture~\cite{chen2022simple}, with the encoder and decoder comprising [2, 4, 4, 8] and [2, 2, 2, 2] NAFBlocks per stage, respectively. The model is trained for 200K iterations on NVIDIA RTX 4090 GPUs, with a per-GPU batch size of 16. After training, DaReg is frozen, and the segmentation head is independently trained for 5K iterations.
In Stage 2, we fine-tune Stable Diffusion 2.1~\cite{rombach2022high} using LoRA with a rank of 16, while keeping DaReg and the segmentation head frozen as fixed modules. We employ the AdamW optimizer with a learning rate of 5e-5, a weight decay of 1e-10, and a per-GPU batch size of 4. This stage is trained for 300K iterations on NVIDIA RTX 4090 GPUs.

\noindent\textbf{Evaluation Metrics.} To ensure fair and comprehensive comparison, we employ both perceptual and distortion-based image quality metrics. For full-reference  perceptual evaluation, we use DISTS, which assess perceptual similarity via features from pre-trained deep networks. For no-reference  assessment, we adopt MANIQA, MUSIQ, and CLIPIQA, all trained to predict human-perceived quality. In particular, as these metrics are not HDR-specific, we computed them in the tone-mapped domain using a standard \(\mu\)-law operator.  For distortion-based metrics, we include PU-PSNR and PU-SSIM, specifically tailored for HDR images, through the official implementation  \cite{azimi2021pu21} with a peak luminance of 1000 nits. We compute HDR-VDP2 on linear HDR images with the ``rgb-native'' color encoding. The data is treated as display-referred, scaled to a peak luminance of 1000 nits, with a viewing angle of $30^\circ$ and viewing distance of 0.55 m.

\subsection{Comparison with SOTA Methods}
\noindent\textbf{Compared Methods.} We compare GMODiff against state-of-the-art CNN- and Transformer-based methods specifically designed for multi-exposure HDR image reconstruction, including DeepHDR~\cite{wu2018deep}, AHDRNet~\cite{yan2019attention}, SCTNet~\cite{tel2023alignment}, and HDR-Trans~\cite{liu2022ghost}. In addition, we also evaluate against recent advanced approaches: the diffusion-based DiffHDR~\cite{yan2023toward}, the optical-flow-based SAFNet~\cite{kong2024safnet}, and the deep-unfolding-based AFUNet~\cite{li2025afunet}. For fair comparison, we retrain all these methods on our collected dataset using their official open-source implementations. We further compare with existing LDM-based HDR reconstruction methods~\cite{google2024ultrahdr,guan2025hdr,wang2025lediff}, which leverage pre-trained LDMs and are fine-tuned on significantly larger datasets than ours. Notably, since UltraFusion \cite{google2024ultrahdr} is a multi-exposure fusion method that cannot generate HDR images, we only include it for qualitative comparison.

\begin{figure*}[tb]
\centering
\includegraphics[width=1\linewidth]{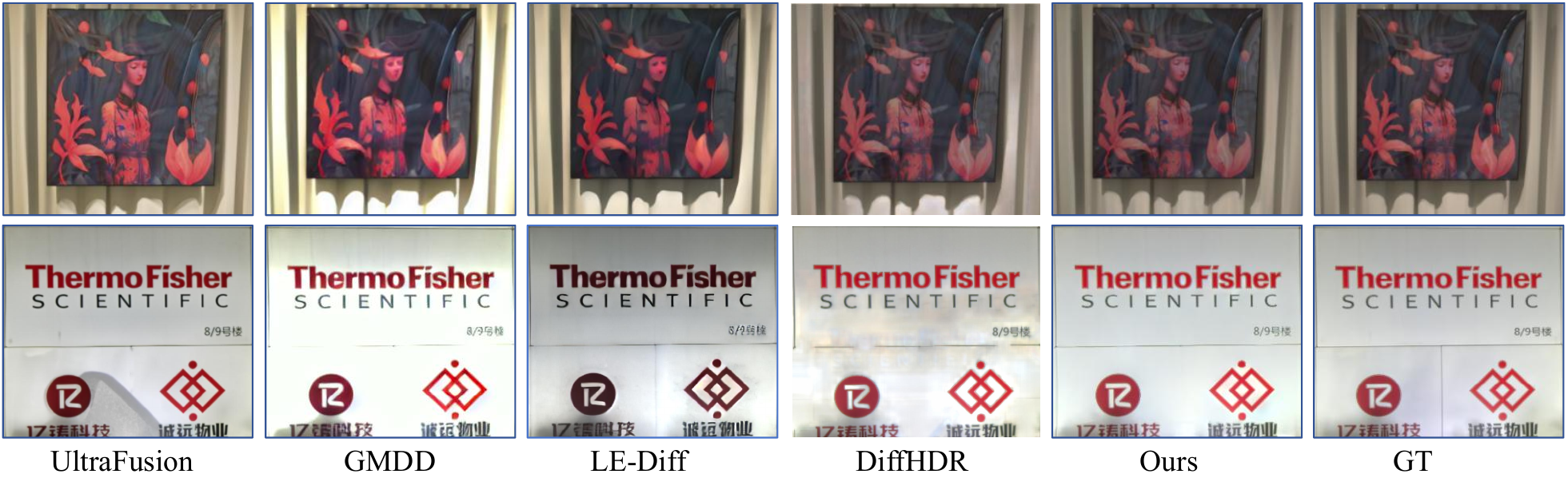}
\caption{Qualitative comparison of HDR reconstruction results with state-of-the-art LDM-based methods. Our GMODiff outperforms existing LDM-based methods in recovering fine details, preserving physical illumination consistency and suppressing artifacts in overexposed/motion regions.
}
\label{compare3}
\end{figure*}
\noindent\textbf{Qualitative Results.}
We present several challenging scenarios. As shown in Fig.~\ref{compare1}, severe camera motion invalidates accurate pixel-wise alignment, leaving only semantic cues for reconstruction. The left case in Fig.~\ref{compare2} involves both human motion and over-saturation, causing severe information loss in the input LDR sequence. In the right example, despite the informative reference frame, camera motion still induces misalignment and degrades reconstruction quality.

As shown in Fig.~\ref{compare1} and \ref{compare2}, existing end-to-end methods consistently suffer from obvious artifacts or ghosting in these challenging regions. Although DiffHDR exploits the generative ability of diffusion models, it still fails to produce visually plausible results due to limited training data. In contrast, our GMODiff combines input cues with powerful priors from pre-trained LDMs to robustly inpaint artifact-prone regions without unrealistic hallucinations, yielding more accurate and perceptually faithful results.
We further compare GMODiff with existing LDM-based methods in Fig. ~\ref{compare3}. GMDD \cite{guan2025hdr} and LE-Diff \cite{wang2025lediff} exhibit severe detail loss because LDMs operate in a compressed latent space. Specifically, GMDD cannot recover missing content in over-exposed areas, while LE-Diff fails to model physical illumination, leading to large brightness and color deviations from the GT. UltraFusion \cite{google2024ultrahdr} is trained solely on LDR images and thus cannot generate real HDR outputs; it also relies heavily on optical-flow alignment, producing obvious artifacts when alignment fails. By contrast, guided by the regression module, GMODiff strictly follows physical constraints and avoids latent-compression information loss via the degradation-aware decoder, generating more realistic and visually pleasing results.

Qualitative comparisons with Prabhakar \etal \cite{prabhakar2019fast}, Sen \etal \cite{sen2012robust}, and Tursun \etal \cite{Tursun2016data} are provided in the \textbf{supplementary material}.

\noindent\textbf{Quantitative Results.} Quantitative comparisons on publicly available datasets are summarized in Tab.~\ref{tab:main_results} and  Tab.~\ref{tab:main_results2}. 
Our method consistently outperforms competing approaches across a diverse set of perceptual metrics. 
Specifically, it achieves notable gains in both full-reference metrics and no-reference metrics, demonstrating its superior ability to reconstruct visually pleasing and perceptually faithful HDR images.
Compared to AFUNet~\cite{li2025afunet}, the state-of-the-art deep unfold method, our approach attains competitive performance in distortion-based metrics (PU-PSNR and PU-SSIM) while achieving significantly better perceptual quality-highlighting that leveraging regression priors to guide pre-trained latent diffusion models enables an effective trade-off between fidelity and perceptual realism.
Moreover, in contrast to diffusion-based alternatives such as DiffHDR~\cite{yan2023toward}, our method fully exploits the rich image priors embedded in large-scale pre-trained LDMs. 
DiffHDR, trained only on limited in-domain data, lacks robust generic priors and struggles in complex real-world scenarios. Notably, since LE-Diff \cite{wang2025lediff} recalculates physical brightness during the reconstruction process, its HDR results suffer from severe brightness discrepancies, leading to extremely low full-reference metrics. Therefore, we do not report it in the table. The user study is shown in Fig. \ref{us}, with details provided in the supplementary material.

\subsection{Ablation Studies}
\begin{figure}[tb]
\centering
\includegraphics[width=1\linewidth]{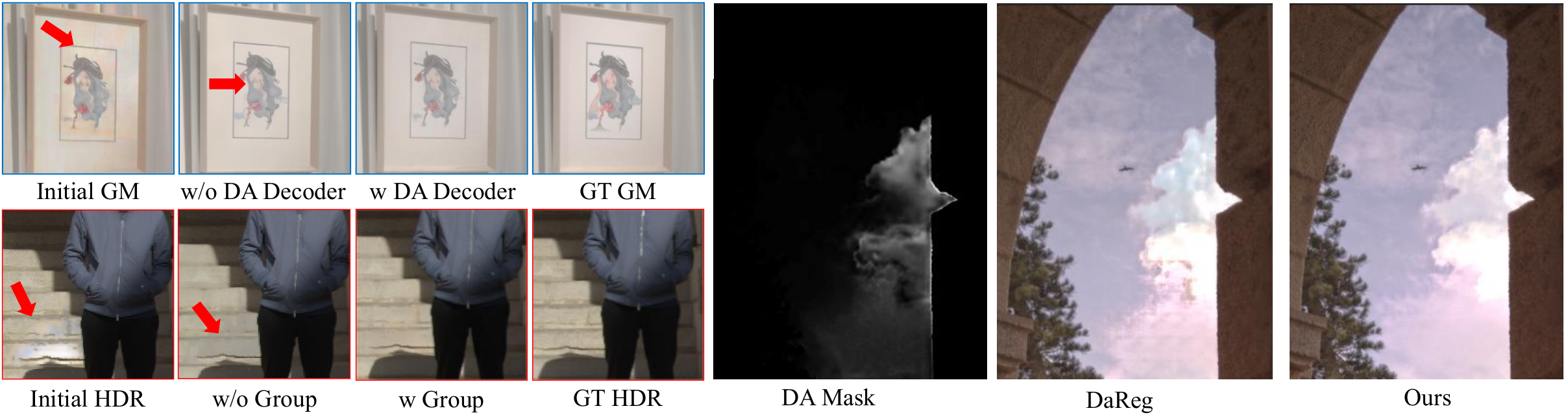}
\caption{
Qualitative results of our ablation study.
}
\label{figab}
\end{figure}

\begin{figure*}[t]
    \begin{minipage}{0.48\linewidth}
        \centering
        \captionsetup{type=table} 
        \Large
        \caption{Ablation study of the each component. The best and second best results are colored with \textcolor{red}{red} and \textcolor{iccvblue}{blue}.}
        \label{tab:ablation} 
        
        \resizebox{1\columnwidth}{!}{
            \begin{tabular}{cccc}
                \toprule[0.15em]
                Type & PSNR$\uparrow$ & M-IQA$\uparrow$ & MUSIQ$\uparrow$  \\
                \midrule[0.15em]
                DaReg &\textcolor{red}{36.87} & 0.5819 & 60.57  \\
                DaReg-Seg &35.47 & 0.5803 & 59.77  \\
                \midrule
                w/o $\mathcal{L}_r$ &36.60 & \textcolor{iccvblue}{0.5875} & \textcolor{iccvblue}{60.84}  \\
                w/o Group Training &\textcolor{iccvblue}{36.76} & 0.5858 & 60.75  \\
                w/o DA Mask &36.49 & 0.5862 & 60.73  \\
                w/o DA Decoder &30.57 & 0.5754 & 59.93  \\
                GMODiff (ours) &{36.68} & \textcolor{red}{0.5881} & \textcolor{red}{60.96}  \\
                \bottomrule[0.15em]
                \label{abb}
            \end{tabular}
        }
    \end{minipage}
    \hfill 
    \begin{minipage}{0.48\linewidth}
        \centering
        \captionsetup{type=figure} 
        \Large
        \includegraphics[width=1\textwidth]{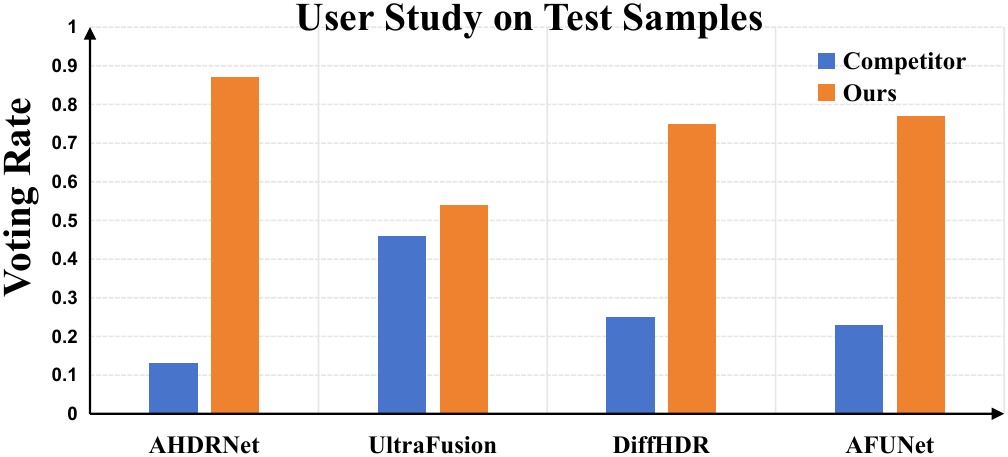}
        \caption{The user study results on the benchmark test set. Details can be found in the supplementary materials.}
        \label{us}
    \end{minipage}
\end{figure*}

To validate the effectiveness of each core component in the proposed GMODiff framework, we conduct comprehensive ablation experiments on the benchmark test set, with quantitative results reported in Tab. \ref{abb} and qualitative results presented in Fig. \ref{figab}. We take the full GMODiff model as the baseline to ablate key modules one by one, and also evaluate the standalone performance of DaReg for comparison. DaReg-Seg refers to the joint training of the reconstruction branch and segmentation head, which leads to a noticeable performance drop. In the second stage of the framework, removing the DA Decoder causes a significant performance decline, as the latent space of LDM fails to preserve the rich detail information contained in the initial GM. Thanks to the rational design of the overall framework, the model only experiences slight performance degradation when other modules are ablated. Compared with the original regression-based DaReg model, GMODiff achieves a favorable trade-off between perceptual quality and distortion metrics with acceptable computational overhead.

\begin{table}[t]
\caption{Inference time comparison on a $1080\times1920$ image.}
\centering
\begin{center}
\vspace{-5mm}
\resizebox{1\columnwidth}{!}{
\begin{tabular}{c|cccccccc}
\toprule[0.15em]
Time (s) $\downarrow$ & AHDR \cite{yan2019attention} & DiffHDR \cite{yan2023toward} & AFUNet \cite{kong2024safnet} & SCTNet \cite{tel2023alignment} & UltraFusion \cite{chen2025ultrafusion} &GMDD \cite{guan2025hdr} &LE-Diff \cite{wang2025lediff}& Ours \\
\midrule[0.15em]
Inference Time & \textcolor{red}{0.24s} & 13.52s & 6.16s & {5.66s} &93.4s &39.2s &132.7s& \textcolor{iccvblue}{0.93s} \\
\bottomrule[0.15em]
\end{tabular}}
\vspace{-3mm}
\label{table:time_comparison}
\end{center}
\end{table}

\subsection{Inference Time Analysis}
Tab.~\ref{table:time_comparison} compares inference latency on $1920 \times 1080$ HDR reconstruction.  
Our method is overall second-fastest, slightly slower than the lightweight CNN-based AHDRNet~\cite{yan2019attention}, and significantly faster than transformer-heavy models (AFUNet~\cite{li2025afunet}, SCTNet~\cite{tel2023alignment}), which, despite using efficient window-based attention, incur high latency due to deep stacking of Transformer blocks and processing at high spatial resolutions.
Among diffusion-based approaches, UltraFusion~\cite{chen2025ultrafusion} requires over a minute per image, owing to patch-wise processing with 50 denoising steps, optical flow alignment, and ControlNet overhead. DiffHDR~\cite{yan2023toward}, despite using only 5 steps, operates in pixel space without latent compression, resulting in high memory bandwidth and slow inference. Since LE-Diff \cite{wang2025lediff} contains two LDMs that separately predict results for different exposures, it takes more than 2 minutes to infer one complete HDR image.
In contrast, our method leverages a pre-trained LDM that performs computation in a highly compressed latent space with a single denoising step, achieving the fastest speed among diffusion-based methods. Moreover, it is compatible with efficient inference frameworks such as \texttt{xFormers}, which further reduce latency through memory, efficient attention and kernel fusion, making it practical for real-world applications.
\section{Conclusion}
We presented GMODiff, a gain map-driven one-step latent diffusion framework for efficient multi-exposure HDR reconstruction. Instead of directly generating HDR radiance in the latent space, GMODiff reformulates the task as degradation-aware gain map refinement within a dual-layer HDR representation, enabling pre-trained LDM priors to be exploited without redesigning the VAE or requiring HDR-specific latent compression. By initializing diffusion from a regression-based gain map estimate and using degradation-aware cues to guide both latent denoising and decoding, the proposed framework combines the structural fidelity of regression models with the perceptual restoration ability of diffusion models while reducing hallucinations in challenging motion and saturated regions. Experiments on benchmark and cross-dataset settings show that GMODiff achieves a favorable balance between distortion fidelity and perceptual quality, with clear improvements in visual realism and artifact suppression.

\section*{Acknowledgment}
This work is supported by NSFC of China under Grant 62301432 and Grant 62306240, the Natural
Science Basic Research Program of Shaanxi under Grant 2023-JC-QN-0685 and Grant QCYRCXM-2023-057, the Fundamental Research
Funds for Central Universities, and Guangdong Basic and Applied Basic Research Foundation 2025A1515011119, the Innovation Foundation for Doctor Dissertation of Northwestern Polytechnical University ZX2025019.

%
%
\bibliographystyle{splncs04}
\bibliography{main}
\end{document}